\documentclass[12pt]{l4dc2020}

\newcommand{\R}{\mathbb{R}}

\newcommand{\pol}{\pi} 
\newcommand{\polparam}{\mathbf{\theta}} 
\newcommand{\polpar}{\pol_\polparam}
\newcommand{\wts}{\mathbf{w}}

\newcommand{\ctre}{u} 
\newcommand{\statee}{s} 

\newcommand{\params}{\mathbf{p}}

\newcommand{\ctr}{\mathbf{\ctre}}
\newcommand{\state}{\mathbf{\statee}}

\newcommand{\hor}{T} 
\newcommand{\traj}{\tau} 

\newcommand{\cst}{c} 

\newcommand{\probb}{p} 
\newcommand{\Ex}{\mathbb{E}} 
\newcommand{\gdis}{q}

\newcommand{\bregdiv}{\phi}
\newcommand{\Lag}{\mathcal{L}}
\newcommand{\lagc}{\mathbf{\lambda}}

\newcommand{\dyn}{\mathbf{f}}
\newcommand{\transpp}{\prime}

\newcommand{\ent}{\mathcal{H}}

\newcommand{\mqi}{\mathbf{\mu}^{q_i}}
\newcommand{\covqi}{\Sigma^{q_i}}
\newcommand{\mdl}{\mathbf{\mu}^\pol}
\newcommand{\covdl}{\Sigma^\pol}

\newcommand{\nor}{\mathcal{N}}
\newcommand{\Nsamp}{N}
\newcommand{\Ntraj}{M}

\newcommand{\Nr}{N_d}
\newcommand{\p}{p}
\newcommand{\Dpe}{g}

\newcommand{\De}{\mathbf{d}}

\title[GPS  Based Control of a High Dimensional AM Process ]{Guided Policy Search Based Control of a High Dimensional Advanced Manufacturing Process}
\usepackage{times}

\author{%
 \Name{Amit Surana} \Email{suranaa@utrc.utc.com}\\
 \Name{Kishore Reddy} \Email{reddykk@utrc.utc.com}\\
 \Name{Matthew Siopis} \Email{siopismj@utrc.utc.com}\\
 \addr Raytheon Technologies Research Center, 411 Silver Lane, East Hartford, CT, 06118, USA%
}

\begin{document}

\maketitle

\begin{abstract}%
In this paper we apply guided policy search (GPS) based reinforcement learning framework for a high dimensional optimal control problem arising in an additive manufacturing process. The problem comprises of controlling the process parameters so that layer-wise deposition of material leads to desired geometric characteristics of the resulting part surface while minimizing the material deposited. A realistic simulation model of the deposition process along with carefully selected set of guiding distributions generated based on iterative Linear Quadratic Regulator is used to train a neural network policy using GPS. A closed loop control based on the trained policy and in-situ measurement of the deposition profile is tested experimentally, and shows promising performance.
\end{abstract}

\begin{keywords}%
Reinforcement learning, Direct policy search, Additive manufacturing control
\end{keywords}

\section{Introduction}
In this paper we apply guided policy search (GPS) a deep reinforcement learning (RL) approach for a high dimensional optimal control problem in an additive manufacturing (AM) process. Deep RL has recently shown unprecedented success in dealing with high dimensional problems \cite{mnih2015human,mnih2016asynchronous,silver2018general,lillicrap2015continuous,DuanCHSA16}. Guided policy search (GPS) \cite{DBLP:conf/icml/LevineK13,icml2014c2_levine14,NIPS2014_5444,levine2015end} is a direct policy search method which uses a neural network (NN) to parameterize the policy, and transforms the policy search into a supervised learning problem, where the training set (which “guides” the policy search to regions of high reward) is generated by simple trajectory-centric algorithms. It employees a regularized importance sampled policy optimization to enable stable and efficient training.

The problem considered in this paper comprises of controlling the process parameters during cold spray (CS), a metal AM process \cite{yin2018cold}. Specifically, the goal is to control the spray nozzle motion, so that layer-wise deposition of material leads to desired geometric characteristics of the resulting part surface while minimizing the material deposited. The relationship between the surface profile growth and nozzle motion is governed by a time dependent nonlinear partial differential equation (PDE). We discretize the PDE in space/time to obtain a finite dimensional nonlinear input-output system, and use a nonlinear optimization framework to calibrate the CS model parameters based on the experimental data. The NN policy is trained using GPS on the calibrated model using a carefully selected set of guiding distributions generated based on iterative Linear Quadratic Regulator. In order to ensure that the errors made by NN controller during layer deposition do not propagate to successive layers, we use suitable domain randomization. Closed loop control based on the trained NN policy was implemented in a lab setup with feedback from laser displacement sensor which measures the surface profile in real time during deposition. The experimental studies show a material saving of upto $35\%$ compared to the state-of-the-art methods used for process parameter selection in the CS applications.

\paragraph{Related Work:} Currently, the process parameter optimization in AM is done offline which does not account for in process uncertainties/disturbances. The need for in-situ monitoring, machine learning and closed loop control has been recognized as a key enabler to improve process repeatability/reliability  \cite{tapia2014review,everton2016review,qi2019applying}. Specifically, in the CS context, \cite{chakraborty2017optimal} apply a model predictive control approach in simulations. We use a similar model/setup but employ a RL framework, and provide an experimental demonstration.

\section{Cold Spray Control Problem}\label{sec:cscontrol}
Cold spray (CS) is an AM process in which powder particles (typically 10 to 50 micron) are accelerated to very high velocities (200 to 1000 m/s) by a supersonic compressed gas jet at temperatures well below their melting point and then deposited on the surface to build new parts or repair structural damages such as cracks and unwanted indentations \cite{yin2018cold}. CS is a multi-scale process with complex physics \cite{wang2015characterization}, making it challenging to optimize the process parameters which result in desired part properties. Additionally, the nozzle motion relative to part/defect geometry needs to be programmed to achieve desired surface deposition profile. Currently, this is accomplished by manual trial and error per part, and often results in a conservative solution leading to excess material deposition which needs to be eventually machined out to meet the required geometric tolerances.  Moreover, the nozzle motion once programmed remains fixed, and is not adapted to account for any process variations/disturbances during the operation. Such an approach of overbuilding could lead to a significant loss of expensive material.
\begin{figure}[hbt!]
  \centering
   \includegraphics[width=0.80\textwidth,trim={0 300 0 0}]{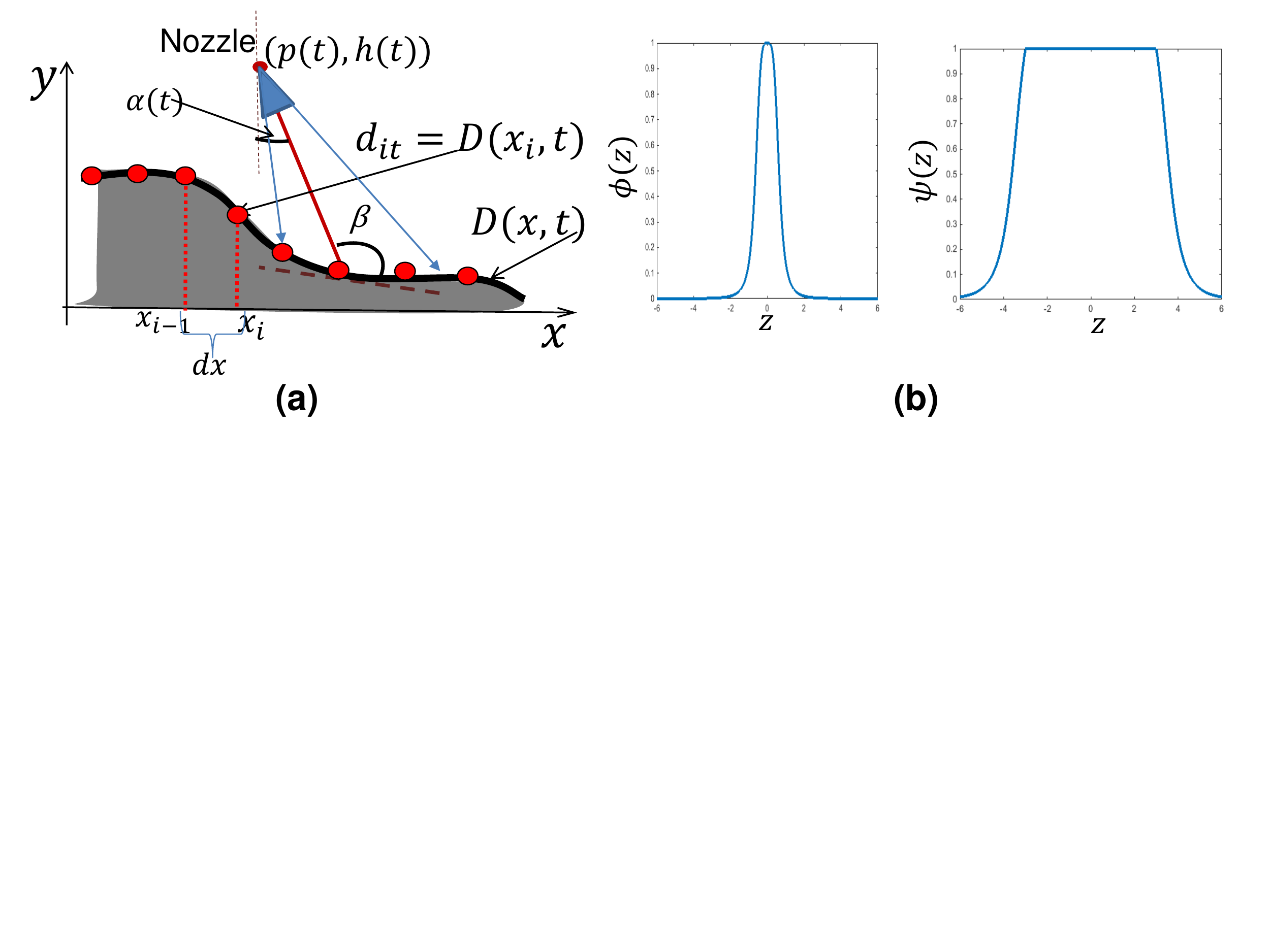}
  \caption{Cold spray model.}
  \label{fig:csmodel}
\end{figure}

The focus of our work is to develop a framework for feedback control of nozzle motion that minimizes the material wastage while achieving desired surface profile. While detailed CS models are highly complex, for our control development purposes it suffices to use a reduced order model described in \cite{chakraborty2017optimal} which captures the impact of nozzle motion/spray behavior on the dynamics of deposition process at a macroscopic scale.  Let $p(t),h(t)$ and $\alpha(t)$ be the spray nozzle's, position and orientation as a function of time, respectively, see Fig. \ref{fig:csmodel}a. Then the surface height profile $D(x,t)$ ($x\in\R$) evolution is governed by a nonlinear integro-differential equation,
\begin{equation}\label{eq:contModel}
\frac{\partial D(x,t)}{\partial t}=\int_{0}^{T}\phi(\tan\beta(t)-\tan\alpha(t))\psi\left(\frac{\tan\beta(t)-\frac{\partial}{\partial x}D(x,t)}{1+\tan\beta(t)\frac{\partial }{\partial x}D(x,t)}\right)dt,
\end{equation}
where, $\tan\beta(t)=\frac{x-p(t)}{h(t)-D(x,t)}$ and the functions
\begin{equation}\label{eq:phidef}
\phi(z)=1-\left(1+\frac{\rho}{z^2}\right)^{-1},\qquad \psi(z)=\frac{1}{c}\left(0.5+\frac{\mbox{atan}\left(-a\max(|z/b|^\kappa,1)\right)}{\pi} \right),\notag
\end{equation}
capture the distribution of particles in the spray cone and the nozzle efficiency, respectively (see Fig. \ref{fig:csmodel}b). We will denote by $\params=(\rho,a,b,c,\kappa)^\prime$ as the model parameter vector, where $\prime$ denotes the vector transpose. In above we have restricted to a one-dimensional model, as often in applications the part is rotated while nozzle motion is confined along the axis of rotation, see Sec. \ref{sec:exp}.

The space-time discretization of Eqn. (\ref{eq:contModel}) leads to a discrete time nonlinear input-output system
\begin{equation}\label{eq:dynacs}
\state_{t+1}=\dyn(\state_t,\ctr_t;\params),
\end{equation}
where, $\state_t=(\De_t^\transpp,\p_t,h_t,\alpha_t)^\transpp$ is the system state, $\ctr_t=(v^x_t,v^h_t,\omega_t)^\transpp$ are the control inputs, and
\begin{equation}\label{eq:mapcs}
\dyn(\state_t,\ctr_t)=\left(
                        \begin{array}{c}
                          d_{1t}+ \Dpe_1(\De_t,\p_t,h_t,\alpha_t;\params)dt\\
                          \vdots \\
                          d_{\Nr t}+ \Dpe_{\Nr}(\De_t,\p_t,h_t,\alpha_t;\params)dt \\
                          p_{t}+v^x_t dt \\
                          h_{t}+v^h_t dt \\
                          \alpha_t+\omega_t dt \\
                         \end{array}
                      \right).
\end{equation}
Here, $\De_t=\left(d_{1t} \cdots  d_{\Nr t} \right)^\transpp$ is the vector of discretized surface height profile $D(x,t)$, with $d_{it}$ being the surface height at the spatial location $x_i=x_0+(i-1)dx,i=1,\cdots,\Nr$ and $\Nr$ is the number of discretized cells of size $dx$, see Fig. \ref{fig:csmodel}a. We assume a kinematic model of the nozzle motion, so that the nozzle position/orientation can be controlled by changing its linear $v^x_t,v^h_t$ and angular $\omega_t$ speeds, respectively. $\Dpe_i$ represent the discretization of the integral term appearing in the Eqn. (\ref{eq:contModel}).

Given initial condition $\state_0=(\De_0^\transpp,p_0,h_0,\alpha_0)^\transpp$, the objective is to determine control sequence $\ctr_t=(v^x_t,v^h_t,\omega_t)^\transpp,t=1\cdots,T$ such that the final surface profile matches a desired profile $\De_f$,
\begin{equation}\label{eq:csctrprob}
\min_{\ctr_1,\cdots,\ctr_T,T} \sum_{t=1}^{T-1} \cst(\state_t,\ctr_t)+\cst_e(\state_T),
\end{equation}
where, $\cst_e(\state_T)=||\De_T-\De_f||^2$ is the terminal cost, and $\cst(\state_t,\ctr_t)=w_0+w_1(v^x_t)^2+w_2(v^h_t)^2+w_3\omega_t^2$ is running cost which imposes penalties on nozzle's speed and angular rate, and the total spraying time $T$ which is not known apriori and needs to be optimized as well. Additionally, bounds can be imposed on velocity and angular rates: $v^x_{\mbox{min}}\leq v^x_t\leq v^x_{\mbox{max}}$, $v^h_{\mbox{min}}\leq v^h_t\leq v^h_{\mbox{max}}$ and $\omega_{\mbox{min}}\leq \omega_t\leq \omega_{\mbox{max}}$, respectively. This optimal control problem can be solved using model predictive control (MPC), see \cite{chakraborty2017optimal}. However, since typically $\Nr\gg1$ to resolve the surface profile sufficiently, such an approach can be computationally demanding for real time deployment. We explore direct policy search based RL methods which are well suited for such high-dimensional control applications, since they scale gracefully with dimensionality and offer appealing convergence guarantees \cite{peters2008reinforcement,kober2013reinforcement}.  However, it is often necessary to carefully choose a specialized policy class to learn the policy in a reasonable number of iterations without getting trapped into poor local optima.

\section{Guided Policy Search}
Guided policy search (GPS) is a recently proposed direct policy search method which uses a neural network (NN) to parameterize the policy and transforms the policy search into a supervised learning problem, where the training set (which “guides” the policy search to regions of high reward) is generated by simple trajectory-centric algorithms. NN provide a general and flexible representation which can represent a broad range of behaviors. However, naive supervised learning will often fail to produce a good policy and a regularized importance sampled policy optimization has been proposed. We will use the end-to-end GPS formulation \cite{levine2015end}, which we briefly review. Consider a finite horizon stochastic optimal control problem
\begin{equation}\label{eq:minprob1}
\min_{\polpar}\Ex_{\polpar}[\sum_{t=1}^\hor \cst(\state_t,\ctr_t)],
\end{equation}
where, $\cst(\state,\ctr)$ is the cost function, and the expectation $\Ex$ is taken under the randomized policy $\pol_{\polparam}(\ctr_t|\state_t)$ which is parameterized by $\polparam$ and the uncertain system dynamics with state transition probability $\probb(\state_{t+1}|\state_t,\ctr_t)$. Let $\gdis(\ctr_t|\state_t)$ be a guiding policy, then the problem (\ref{eq:minprob1}) can be reformulated as an equivalent problem:
\begin{eqnarray}\label{eq:minprob2}
&&\min_{\gdis, \polpar}\Ex_{\gdis}[\sum_{t=1}^\hor \cst(\state_t,\ctr_t)],\quad \gdis(\ctr_t|\state_t)=\pol_{\polparam}(\ctr_t|\state_t),\quad \forall t, \state_t,\ctr_t.
\end{eqnarray}
which has an infinite number of constraints. To make the problem tractable one can use the moment matching approach, e.g.
\begin{eqnarray}\label{eq:minprob}
&&\min_{\gdis, \polpar}\Ex_{\gdis}[\sum_{t=1}^\hor \cst(\state_t,\ctr_t)],\quad \Ex_{\gdis(\ctr_t|\state_t)\gdis(\state_t)}[\ctr_t]=E_{\pol_{\polparam}(\ctr_t|\state_t)\gdis(\state_t)}[\ctr_t],\quad \forall t.
\end{eqnarray}
This constrained problem can be solved by a dual descent method, e.g. Bregman Alternative Direction Method of Multipliers (BADMM) Lagrangian formulation leads to
\begin{eqnarray}
\Lag(\gdis,\polpar)&=&\sum_{t=1}^\hor \left[ \Ex_{\gdis(\state_t,\ctr_t)}(\cst(\state_t,\ctr_t))+\lagc_{\mu t}^{\transpp}(\Ex_{\polpar(\ctr_t|\state_t)\gdis(\state_t)}[\ctr_t]
-\Ex_{\gdis(\state_t,\ctr_t)}[\ctr_t])+\nu_tD_{KL}(\gdis,\polpar)\right],\notag
\end{eqnarray}
where we have taken the Bregmann divergence to be the KL divergence $D_{KL}(\polpar|\gdis)$. This leads to following iterative optimization scheme,
\begin{eqnarray}
\gdis &\leftarrow & \arg\min_{\gdis} \sum_{t=1}^\hor \left[ \Ex_{\gdis(\state_t,\ctr_t)}[\cst(\state_t,\ctr_t)-\lagc_{\mu t}^{\transpp}\ctr_t]+\nu_t\bregdiv_t(\gdis,\polparam)\right],\label{eq:BADMMiterq}\\
\polparam & \leftarrow & \arg\min_{\polparam}\sum_{t=1}^\hor \left[\Ex_{\polpar(\ctr_t|\state_t)\gdis(\state_t)}[\lagc_{\mu t}^{\transpp}\ctr_t]+\nu_t\bregdiv_t(\polparam,\gdis)\right],\label{eq:BADMMiterpol}\\
\lagc_{\mu t} & \leftarrow& \lagc_{\mu t}+\alpha\nu_t (E_{\pol_{\polparam}(\ctr_t|\state_t)\gdis(\state_t)}[\ctr_t]-\Ex_{\gdis(\state_t,\ctr_t)}[\ctr_t]),\quad t=1,\cdots,\hor, \label{eq:BADMMiterlam}
\end{eqnarray}
Note that for each $t=1,\cdots,\hor$, $\lagc_{\mu t}$ is a vector of Lagrangian multiplier with same dimension as the control $\ctr_t$. $\alpha$ can be chosen in range $[0,1]$, lower values lead to better numerical stability. The weights $\nu_t$ are initialized to low values such as $0.01$ and incremented based on a schedule which adjusts the KL-divergence penalty to keep the policy and trajectory in agreement by roughly the same amount at all time steps. The above steps can be simplified under Gaussian assumptions:
\begin{itemize}
  \item $\polpar(\ctr_t|\state_t)\sim \nor(\ctr_t;\mdl(\state_t;\wts),\covdl)$, where the policy parameters are $\polparam=(\wts,\covdl)$. We will use a neural network  (NN) with weights $\wts$ to represent the mean policy $\mdl(\state_t;\wts)$, while $\covdl$ is policy covariance which is assumed to be state independent,
  \item $\gdis=\sum_{i=1}^\Ntraj \gdis_i$ is mixture of Gaussians with $\gdis_i(\ctr_t |\state_t)\sim \nor(\ctr_t;\mqi_t(\state_t),\covqi_t)$
\end{itemize}
where, $\nor(\cdot,\mu,\Sigma)$ is a multinomial normal probability distribution with mean vector $\mu$ and covariance matrix $\Sigma$.  With this assumption, the overall GPS algorithm is summarized below. Steps 1-4 are repeated for a preselected $K$ iterations or else if a prescribed convergence criterion is met.

\paragraph{Step 1:} Solve optimization (\ref{eq:BADMMiterq}) which under the Gaussian assumption simplifies to
\begin{eqnarray}
\arg\min_{\gdis_i} \sum_{t=1}^\hor \Ex_{\gdis_i(\state_t,\ctr_t)} \left[ \cst_i(\state_t,\ctr_t) -\ent(\gdis_i(\ctr_t|\state_t))\right],\nonumber \label{eq:BADMMiterqisimp}
\end{eqnarray}
where, $\cst_i(\state_t,\ctr_t)=\frac{\cst(\state_t,\ctr_t)}{\nu_t}-\frac{1}{\nu_t}\ctr_t^\transpp \lagc_{\mu t}^i -\log (\polpar(\ctr_t|\state_t))$
and, $\ent$ is the standard differential entropy. A local optimal solution (which determines $\mqi_t,\covqi_t$) of the above problem around a selected nominal trajectory can be obtained via a variation of iterative LQR (iLQR) \cite{li2004iterative}.
The nominal trajectory can be constructed in a variety of ways. e.g. from demonstrations, using randomized control inputs or via solution obtained using MPC \cite{DBLP:journals/corr/ZhangKLA15}. 

\paragraph{Step 2:}  Sample trajectories $\traj^{ji}=\{(\state_t^{ij},\ctr^{ij}_t):t=1:\hor\},j=1,\cdots,\Nsamp$ from
\begin{equation}\label{eq:tarjprob}
\gdis_i(\{\state_1,\ctr_1,\cdots \state_T,\ctr_T \})=\probb_{i1}(\state_1)\prod_{t=1}^\hor\gdis_i(\ctr_t|\state_t)\probb(\state_{t+1}|\state_t,\ctr_t).
\end{equation}
induced by each of the guiding distributions $\gdis_i, i=1,\cdots, \Ntraj$ with $\probb_{i1}(\state_1),i=1,\cdots,\Ntraj$ being the initial state distribution.

\paragraph{Step 3:} Train the NN representing the policy mean $\mdl(\state;\wts)$ with a modified objective (\ref{eq:BADMMiterpol}) which under Gaussian assumption becomes:
\begin{equation}\label{eq:DLcost}
\arg\min_{\wts}\frac{1}{2\Nsamp}\sum_{t=1}^\hor\sum_{i=1}^\Ntraj\sum_{j=1}^\Nsamp\left[\nu_t(\mqi_t(\state_t^{ij})-\mdl(\state_t^{ij};\wts))^\transpp(\covqi_t)^{-1}(\mqi_t(\state_t^{ij})-\mdl(\state_t^{ij};\wts))+2\lagc_{\mu t}^{i\transpp}\mdl(\state_t^{ij};\wts)\right].
\end{equation}
The policy covariance $\covdl$ can be computed directly as $\covdl=\left[\frac{1}{\Ntraj \hor} \sum_{i=1}^\Ntraj \sum_{t=1}^\hor (\covqi_t)^{-1}\right]^{-1}$.


\paragraph{Step 4:} Update the Lagrange multipliers
\begin{equation}\label{eq:lagupdate}
\lagc_{\mu t}^i \leftarrow \lagc_{\mu t}^i+\alpha\nu_t \frac{1}{\Nsamp} \sum_{j=1}^\Nsamp [\mdl(\state_t^{ij};\wts)-\mqi_t(\state_t^{ij})],\quad t=1,\cdots,\hor; \quad i=1,\cdots,\Ntraj.
\end{equation}

\subsection{Adaptation of GPS Framework}\label{sec:adpathGPS}
\begin{figure}[h!]
  \centering
   \includegraphics[width=0.7\textwidth,trim={0 300 0 0}]{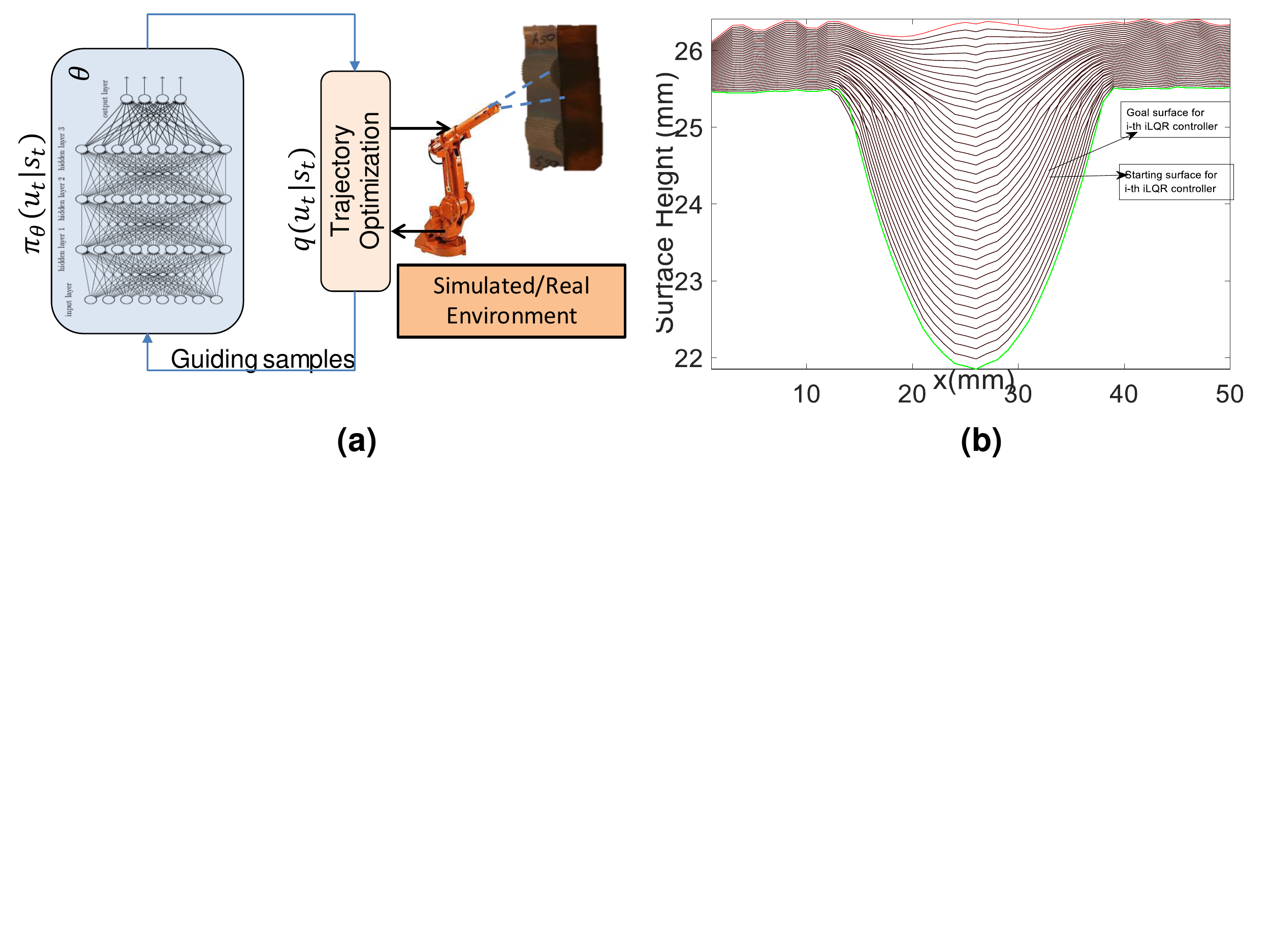}
  \caption{(a) GPS Schematic. (b) Intermediate goals used for generating guiding distributions.}
  \label{fig:arch}
\end{figure}
Application of GPS framework to the CS control problem stated in Section \ref{sec:cscontrol} required further adaptations. For the selected CS application, achieving the desired profile required multiple passes. Hence, intermediate goals were defined for computing iLQR based local optimal controllers which will depend on part/defect geometry. Figure \ref{fig:arch}b  shows examples of intermediate goal profiles $\De_f^i$ for the test part considered in our work (see Section \ref{sec:setup}) which were determined so that in each pass least amount of build up happens on the flat portion and maximum deposition occurs in the notch. The objective for the $i-$th iLQR was to achieve the desired goal profile $\De_f^i$ starting from the previous goal $\De_f^{i-1}$. Furthermore, the objective function in (\ref{eq:csctrprob}) was modified for each $i-$th iLQR with this goal profile, and the terminal cost term $\cst_e$ was absorbed in the total cost leading to  $\sum_{t=1}^{T}( \cst(\state_t,\ctr_t)+||\De_t-\De_f^i||^2)$. In order to generate training data for NN, initial starting surface for each iLQR was perturbed which not only involved randomized to account for measurement noise in the experiments \cite{tobin2017domain}, but also included the propagation effect of previous iLQR controller in achieving their goal surface profiles. Generation of rich training sets which covers different possibilities of intermediate states is necessary in sequential decision making tasks: as if NN control actions resulted in surface buildups never encountered in the training, that will further drive the NN to make errors resulting in cascading failure. We will refer to NN controller trained using GPS as GPS based controller (GPSC).

\section{Results}\label{sec:exp}

\subsection{Experimental Setup}\label{sec:setup}
\begin{figure}[hbt!]
  \centering
   \includegraphics[width=0.70\textwidth,trim={0 170 0 0}]{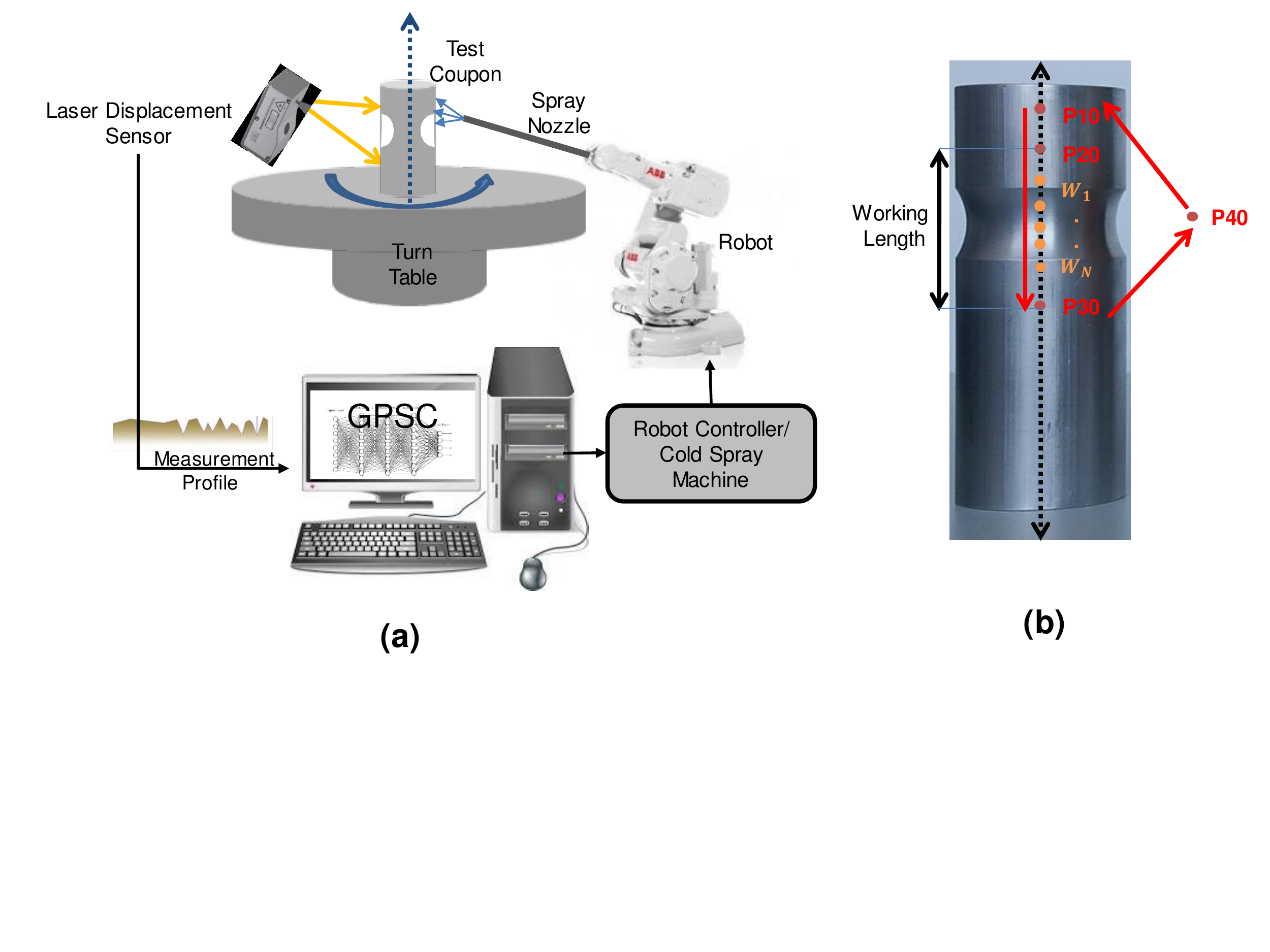}
  \caption{a) Cold spray closed loop experimental setup. b) Cylindrical test coupon with a notch.}
  \label{fig:setup}
\end{figure}

The control setup for GPSC demonstration comprised of (see Fig. \ref{fig:setup}): CS machine/robot,  laser displacement sensor (LDS), and standard PC to host the GPSC algorithm.  The CS machine comprised of gas heater, powder feeder, and nozzle mounted on a 5-axis robot. A turn table was used to rotate the part to be coated, and thus requiring only a linear motion of nozzle for deposition on the complete surface of an axis symmetric part. The LDS had a $640$ pixel resolution with a sampling rate of $1$kHz. LABVIEW was used for implementing the control logic which sequentially activated the LDS to provide surface profile measurements, followed by the GPSC code for nozzle speed computation based on the measurements, and then transmitted the nozzle speed commands to the robot controller. For testing the GPSC framework, we used a aluminium cylindrical test coupon with a symmetrical notch representing a defect to be repaired as shown in Fig. \ref{fig:setup} b.

\subsection{Model Calibration}\label{sec:calib}
\begin{figure}[h!]
  \centering
   \includegraphics[width=0.9\textwidth,trim={0 340 0 0}]{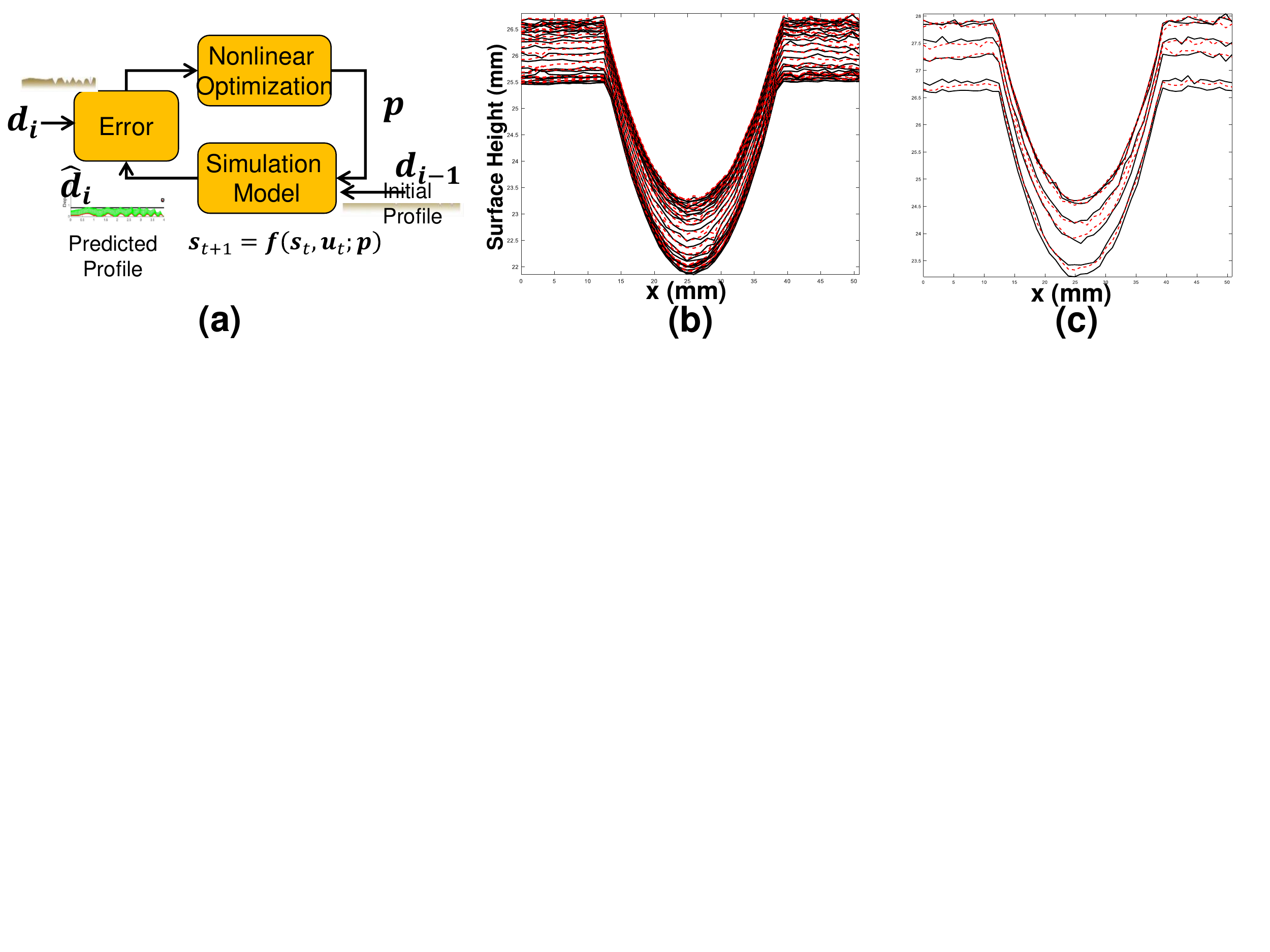}
  \caption{a) Model calibration procedure schematic; b) Calibration results on training data; and c) Calibration results on validation dataset.}
  \label{fig:modelcal}
\end{figure}

We used an optimization framework for CS model parameter calibration. The calibration data was generated by making several consecutive passes on the test coupon, where for each pass the speed was held constant taking values $v^c_i\in \{1,2,3,4,5\}$mm/sec. The surface profiles after each pass were recorded using the LDS, which we denote by $\De_i,i=0,\cdots,N_c$. Let $\hat{\De}_i(\params;\De_{i-1})$ be predicted surface using the CS model (\ref{eq:mapcs}) starting with $\De_{i-1}$, and using parameters $\params$ and constant nozzle speed $\{v^x_{it}=v^c\}_{t=0}^{T_i}$ over the pass. The optimization problem can be defined as (see Fig. \ref{fig:modelcal}a)
\begin{equation}\label{eq:opt}
\min_{\params} \sum_{i=1}^{N_c}||\hat{\De}_i(\params;\De_{i-1})-\De_i||^2.
\end{equation}
We additionally introduced linear constraints $\params_{min}\leq \params\leq \params_{max}$ to bound the parameter values in a desired range. The optimization (\ref{eq:opt}) was solved using fmincon nonlinear optimization routine in MATLAB using a subset of training data, and the remaining set was used for validation. Figure \ref{fig:modelcal}b-c shows the calibration results on training and the validation dataset, respectively. The black curves represents experimental data while dotted red curves show calibrated model predictions.

\subsection{GPSC Deployment}\label{sec:train}
We used the GPSC framework described in Section \ref{sec:adpathGPS}, where only the nozzle speed $v^x_t$ was the controlled variable, the nozzle height $h_t$ and orientation $\alpha_t$ were kept fixed. We used $N_d=100$ for the discretized representation of the surface (see Section \ref{sec:setup}). Consequently, the input layer of the NN consisted of $101$ neurons (with one additional input for the nozzle position), and output layer had one neuron corresponding to the nozzle speed. We found that $1$ hidden layer with $10$ neurons was sufficient for our application. In order to ensure that the output speed remains bounded in range $[0\quad 5]$, we selected a sigmoid activation function for the output layer. The closed loop control was implemented at every pass, i.e. after measurement from LDS was available the controller computed nozzle speed for the entire next pass by propagating the calibrated CS model. This setting of feedforward/feedback control is expected to commonly arise in AM applications, where meaningful sensing data may be only available after one or many layers have been deposited. For repetitive spraying, a cyclic robot motion plan was created, which consisted of the two sets of waypoints (see Fig. \ref{fig:setup} b): $P10,P20,P30,P40$ are the teach waypoints and remain fixed, while  the control waypoints $W_1,\cdots,W_N$ depended on the nozzle speed control commands generated by GPSC.


\subsection{Findings}

\begin{figure}[h!]
  \centering
   \includegraphics[width=0.60\textwidth,trim={0 30 0 10}]{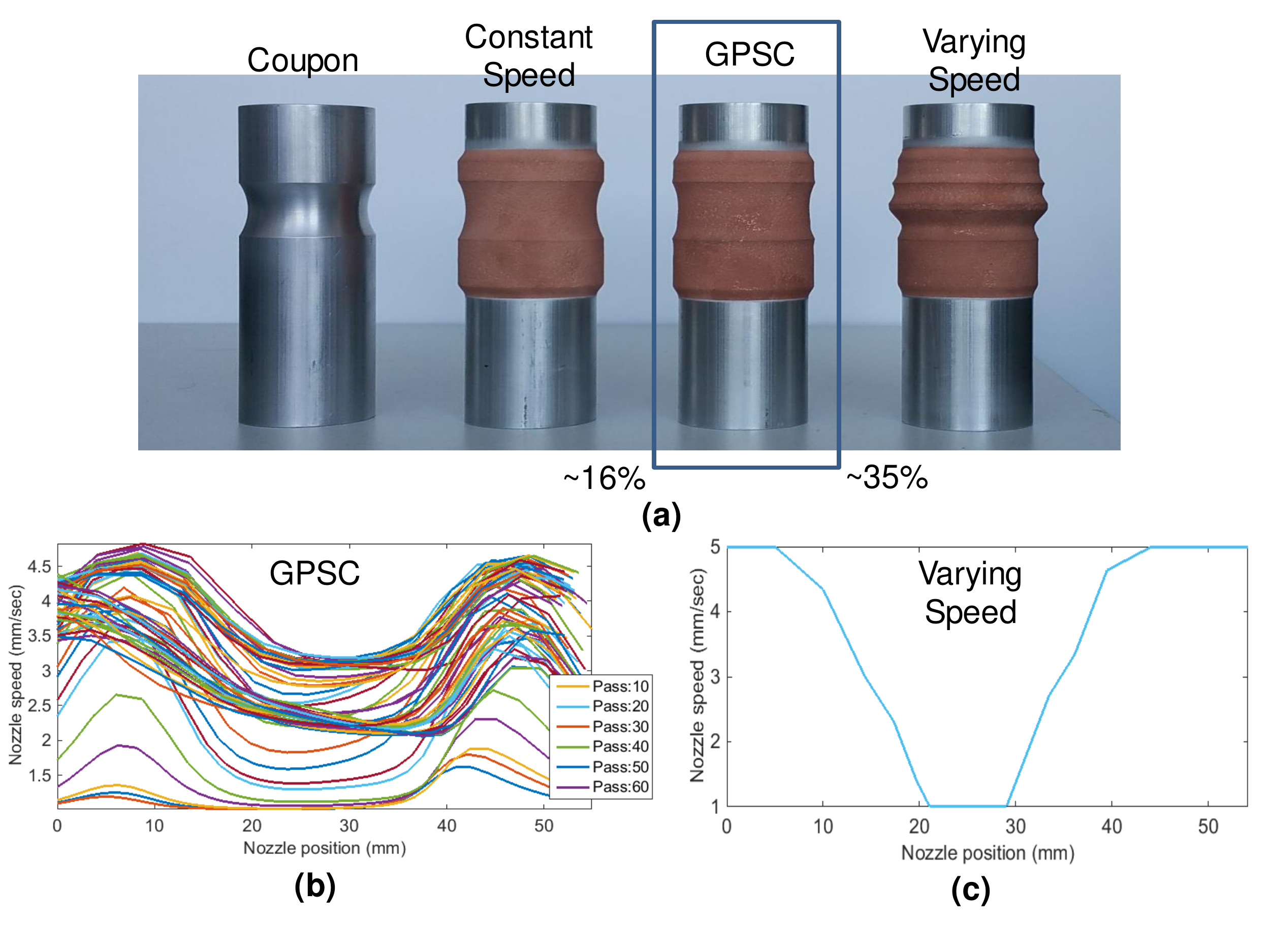}
  \caption{Comparison of different approaches. a) Test coupon after coating, b) Speed profile computed based on GPSC during different passes, c) Hand crafted varying speed profile.}
  \label{fig:results}
\end{figure}
We compare performance of GPSC with two open loop control strategies: constant speed and varying speed. In both these cases the nozzle speed is not changed based on the in-situ LDS measurements. For the constant speed case the speed of nozzle is set to be a constant value of $1$mm/sec. For varying speed case the speed changes with nozzle location to prescribed values as shown in Fig. \ref{fig:results}c. This speed profile was hand crafted by expert so that nozzle moves fastest in flat portion of coupon resulting in minimal deposition and moves slower in the notch area which requires more deposition. The nozzle speed profiles generated for the different passes during GPSC is shown in the Fig. \ref{fig:results}b. Note that GPSC speed profiles are initially similar to the varying speed profile, but adapt as more layers were build to account for various sources of uncertainties/disturbances including robots precision in following the commanded speed, wobbling of the cylinder during rotation, and the variations in powder characteristics. Due to lack of this adaptation in the open loop  varying speed case, the final deposit was significantly irregular (see Figure \ref{fig:results}a). In fact, GPSC saves $~16\%$ scrap material over constant speed, and around $~35\%$ over varying speed profile.

\section{Conclusions}
In this paper we applied guided policy search based RL approach for a high dimensional optimal control problem arising in cold spray manufacturing process. The approach was experimentally validated and showed promising performance. In future, we plan to explore hierarchical RL framework \cite{barto2003recent,vezhnevets2017feudal,nachum2018data} for automatically learning hierarchical structure in additive manufacturing control problems which has a potential to further improve generalization and transfer to different part geometries.

\acks{Funding provided by UTRC is greatly appreciated.}

\newpage
\bibliography{drl,references}

\end{document}